\documentclass[conference]{IEEEtran}
\IEEEoverridecommandlockouts
\usepackage{cite}
\usepackage{amsmath,amssymb,amsfonts}
\usepackage{algorithmic}
\usepackage{graphicx}
\usepackage{textcomp}
\usepackage[table]{xcolor}
\usepackage[T1]{fontenc}
\usepackage{times}
\ifdefined\pdfobjcompresslevel
\fi
\ifdefined\pdfminorversion
\fi
\usepackage{subcaption} 
\usepackage{colortbl}
\usepackage{multirow}
\usepackage{booktabs}      

\def\BibTeX{{\rm B\kern-.05em{\sc i\kern-.025em b}\kern-.08em
    T\kern-.1667em\lower.7ex\hbox{E}\kern-.125emX}}
\begin{document}

\title{SepPrune: A Separator-based Pruning Framework for Efficient Multimodal Large Language Models}

\author{
Yuchen Wang\textsuperscript{1,2}
\hspace{1.2em}
Qihui Zhu\textsuperscript{1}
\hspace{1.2em}
Yang Liu\textsuperscript{1,3,*}
\hspace{1.2em}
Xiaoyan Sun\textsuperscript{1,2,*}
\hspace{1.2em}
Siying Wu\textsuperscript{1,2,*}
\\[0.4em]
\textsuperscript{1}MoE Key Lab of BIPC, NEL-BITA, University of Science and Technology of China, Hefei, China
\\
\textsuperscript{2}APKL of BIIP, IAI, Hefei Comprehensive National Science Center,
Hefei, China
\\
\textsuperscript{3}ChangXin Memory Technologies, Hefei, China
\\[0.4em]
\{wangyuc, qh333666, lqly\}@mail.ustc.edu.cn,
wsy315@iai.ustc.edu.cn,
sunxiaoyan@ustc.edu.cn
}
\maketitle
\begingroup
\renewcommand\thefootnote{*}
\footnotetext{Corresponding authors.}
\endgroup
\begin{abstract}
Owing to the any-res vision encoder structure, vision tokens in the latest Multimodal Large Language Models (MLMMs, Qwen2.5-VL and InternVL3) are more than before, hindering efficiency and widespread applications. Although many vision token pruning methods have been proposed to alleviate this issue, none of them are entirely satisfactory. Text-dependent approaches that rely on cross-modal attention lack flexibility, cannot pre-prune before the prefill stage, while diversity-based methods incur substantial computational overhead. 
When analyzing the attention scores of vision and text tokens, we observed peaks at modality separator tokens (special tokens that separate visual and textual modalities), which implies separator tokens act as a bridge between different modalities. 
Inspired by this observation, we propose SepPrune, an efficient and plug-and-play vision token pruning method, which uses the separator token as a unified query to evaluate and select the most informative vision tokens. By reusing built-in parameters of the LLM itself as a projection matrix, our method prunes vision tokens without altering the LLM architecture. 
Experiments on widely used Qwen2.5-VL-7B demonstrate that our approach achieves state-of-the-art  results, retaining 96.3\% of the original accuracy when dropping 80.2\% of vision tokens. 

\end{abstract}

\begin{IEEEkeywords}
MLLM, Separator token
\end{IEEEkeywords}

\section{Introduction}
\label{sec:intro}

In recent years, the remarkable success of Large Language Models (LLMs)~ \cite{achiam2023gpt} has propelled the field of artificial intelligence forward. This progress has led to the emergence of Multimodal Large Language Models (MLLMs)~ \cite{bai2025qwen2, zhu2025internvl3,liu2024improved}, which are capable of processing both text and visual information simultaneously.  MLLMs have demonstrated powerful multimodal perception and cognitive capabilities in tasks like Visual Question Answering (VQA) and image captioning. 

However, the introduction of the visual modality also presents significant  challenges. When processing high-resolution visual inputs, such as 4K/8K ultra-high-definition videos, high-megapixel photos from digital cameras, or detailed aerial imagery captured by drones, the vision encoder generates a vast number of vision tokens\cite{DBLP:conf/nips/DongZZCWOZDZLYG24}.  The massive number of tokens generated from visual inputs, far exceeding that of text, imposes substantial computational overhead and significant latency on the model's inference process. Prior research \cite{DBLP:conf/nips/RaoZLLZH21} has indicated that a high degree of redundancy exists within these vision tokens. Consequently, to effectively reduce the inference costs of MLLMs, recent works \cite{chen2024image, zhang2024cls, shang2024llava, bolya2022token,Wang_Li_Yin_Liu_Wang_2026,yin2025liftingveilvisualinformation}  have focused on improving efficiency by pruning redundant vision tokens. 
\begin{figure}[htbp]
    \centering
    \begin{subfigure}[b]{0.45\textwidth}
        \includegraphics[width=\linewidth]{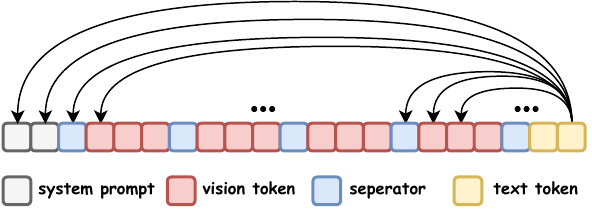}
        \caption{Input Sequences in LLM}
        \label{fig:subfig1}
    \end{subfigure}
    \hfill
    \begin{subfigure}[b]{0.45\textwidth}
        \includegraphics[width=\linewidth]{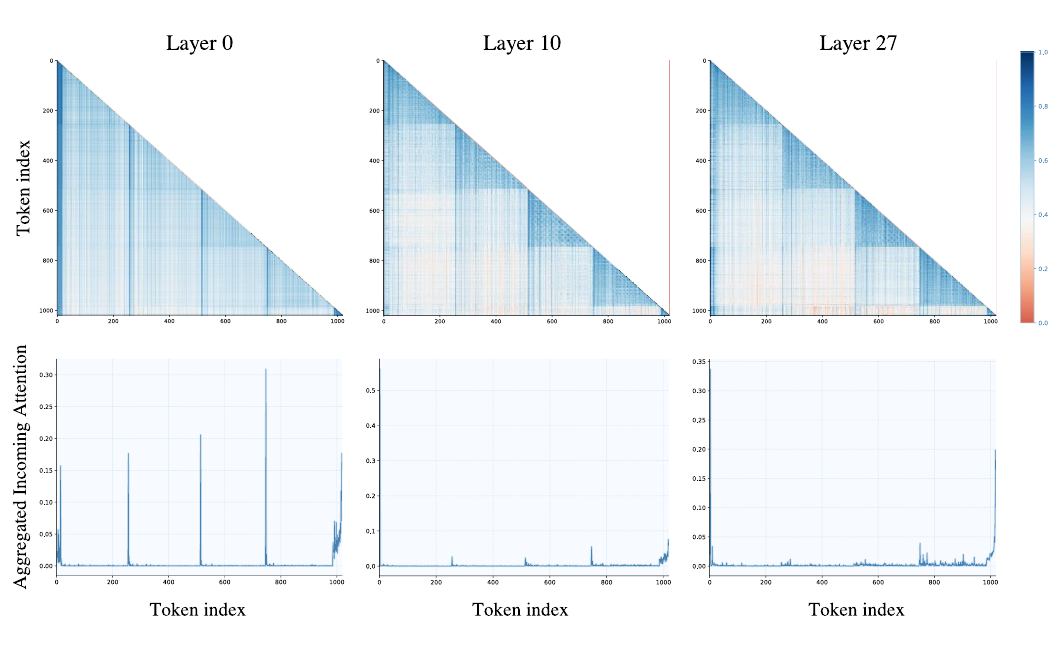}
        \caption{Causal Attention Map}
        \label{fig:subfig2}
    \end{subfigure}
    \caption{(a) Illustration of the input sequence structure to the LLM when processing multiple images  in Qwen2.5-VL-7B ;
(b) The visualization for attention scores of different layers 0, 10, and 27, showing the evolution of attention distribution across different tokens in various layers. }
    \label{fig:combined}
\end{figure}

Existing vision token pruning methods primarily fall into two categories: attention-based and diversity-based strategies. The former \cite{chen2024image,zhang2024cls} often conflicts with hardware-efficient operators like FlashAttention \cite{dao2022flashattention,dao2023flashattention2} or relies on architecture-specific designs like \texttt{[CLS]} tokens. The latter \cite{alvar2025divprune, wen2025stoplookingimportanttokens}, while addressing redundancy, typically incurs a quadratic time complexity of $O(N^2)$, which paradoxically increases latency.

To explore a more efficient pruning mechanism, we first conducted an in-depth analysis of the attention distribution within LLM. Using the advanced Qwen2.5-VL-7B \cite{bai2025qwen2} model as a case study, we visualized its attention behavior when processing multi-image inputs. In typical MLLM input sequences, the structure follows: \texttt{[System Prompt] [Separator] [vision tokens] [Separator] [vision tokens] [Separator] [User Text]} (see Fig. 1a). As shown in Fig. 1b, we observed a surprising phenomenon: in the initial layers of  LLM, attention was not primarily focused on semantically rich visual content tokens. Instead, significant attention peaks emerged at the positions of modality separators, between images and between image and text. These peaks gradually diminish with increasing network depth, indicating that separators play a crucial role in early-stage cross-modal interactions. To quantify the importance of separator, we conducted an ablation study by removing equal numbers of separator tokens, vision tokens, and text tokens. As shown in Fig. 2, compared to removing vision or text tokens, removing separators resulted in the most significant performance degradation. The phenomenon strongly suggests that modality separators are not merely boundary markers but function as cross-modal bridges, serving as key hubs that facilitate early-stage alignment and fusion between vision and language representations.

Inspired by this insight, we propose SepPrune, a novel, efficient, and plug-and-play vision token pruning method that exploits the intrinsic bridging role of separators. Specifically, we utilize the separator token as a unified query and leverage the projection parameters from the first LLM layer to compute the attention scores for all vision tokens. Crucially, we exclude positional encodings during this scoring phase to eliminate location bias, ensuring that tokens are ranked solely based on their semantic importance. We then retain the top-$k\%$ most informative tokens according to a predefined pruning ratio. Since all computations in SepPrune are performed prior to the LLM backbone, the method enables seamless compatibility with acceleration techniques such as FlashAttention and eliminates the need for complex KV-cache management.
We validate SepPrune on leading MLLMs including Qwen2.5-VL-7B \cite{bai2025qwen2} and InternVL3 \cite{zhu2025internvl3}, and conduct extensive experiments on multiple standard MLLM benchmarks. Results demonstrate that SepPrune can significantly reduce the number of vision tokens while maintaining  model performance across all tasks. 

In summary, our main contributions are as follows:

\begin{enumerate}
    \item Through attention analysis, we reveal for the first time the role of modality separators as cross-modal bridges in MLLMs, deepening our understanding of multimodal fusion mechanisms within LLM. 
    \item Based on this insight, we introduce SepPrune, a novel, lightweight, training-free, and plug-and-play vision token pruning strategy that can be seamlessly integrated with existing inference acceleration frameworks. 
    \item Extensive experiments on multiple authoritative benchmarks demonstrate that SepPrune can substantially reduce inference costs while preserving strong model performance. 
\end{enumerate}
\begin{figure}[htbp]
    \centering
    \includegraphics[width=\linewidth]{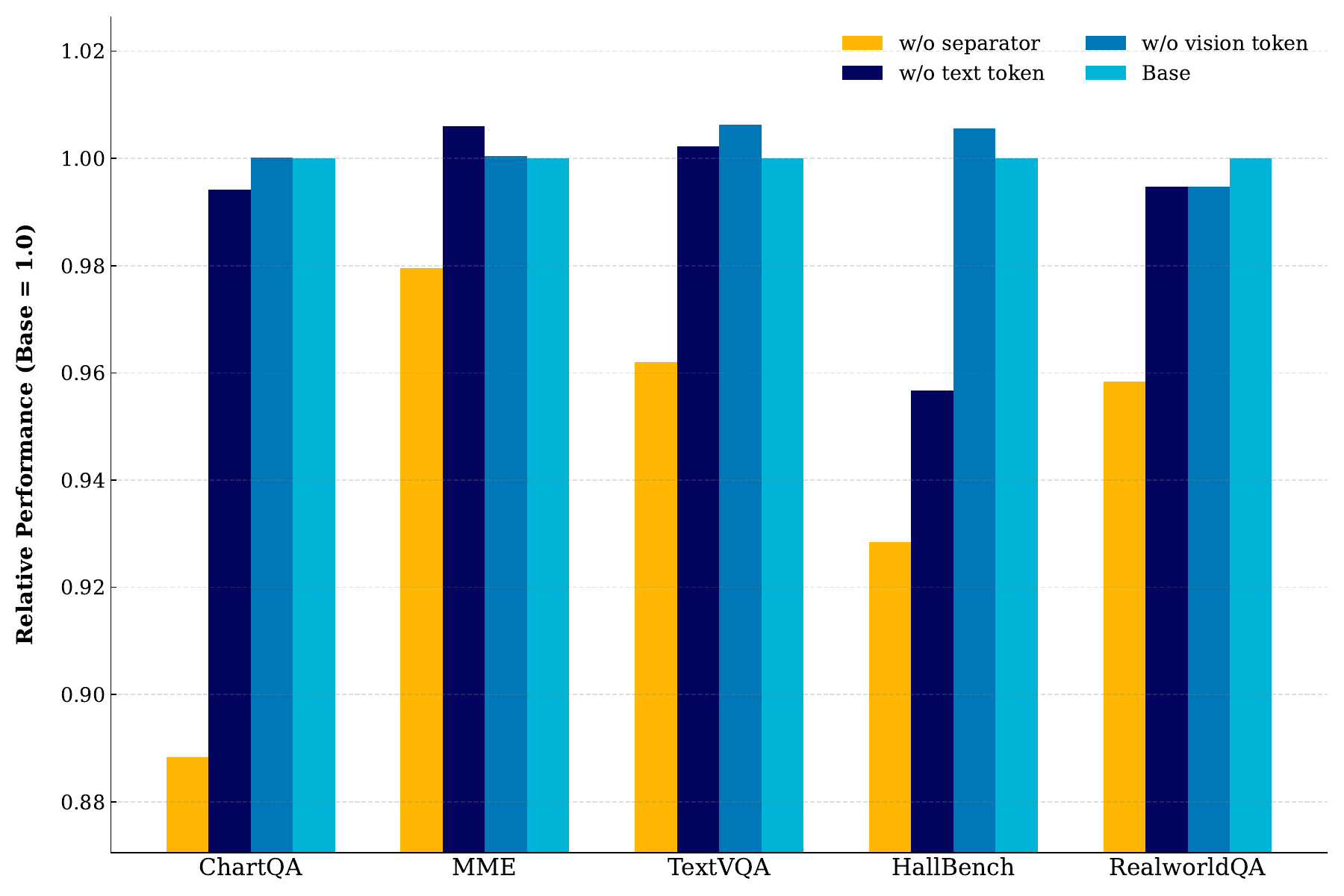}  
    \caption{Ablation study evaluating the impact of separately removing two tokens from each of the separator tokens, vision tokens, and system text tokens within the LLM input sequence. Experiments were conducted on ChartQA, MME, TestQA, HallBench, and RealWorldQA datasets. }
    \label{fig:myfig}
\end{figure}

\section{Related Work}

Vision token pruning aims to enhance inference efficiency by removing redundant tokens. We categorize existing approaches into two streams: attention-based and diversity-based methods.

\subsubsection{Attention-based methods}
These methods evaluate token importance via attention scores. A mainstream paradigm involves utilizing text-vision cross-attention within LLM layers, as seen in FastV \cite{chen2024image}, FitPrune \cite{ye2025fit}, SparseVLM \cite{zhang2025sparsevlmvisualtokensparsification}, MustDrop \cite{liu2024multi}, and PDrop \cite{xing2024pyramiddrop}. However, extracting these scores interrupts the computational flow, precluding the use of efficient attention kernels like FlashAttention \cite{dao2022flashattention}, which limits practical speedups.
Alternatively, methods like HiRED \cite{arif2025hired} and LLaVA-PruMerge \cite{shang2024llava} perform pruning before the LLM by utilizing the \texttt{[CLS]} token as a global query. While avoiding the aforementioned latency issue, their applicability is restricted by the reliance on the \texttt{[CLS]} token, which is absent in many modern vision backbones.

\subsubsection{Diversity-based methods}
These methods focus on reducing information redundancy rather than filtering by importance scores. ToMe \cite{bolya2022token} requires training to dynamically merge similar tokens. Conversely, training-free methods like DivPrune \cite{alvar2025divprune}, CDPruner \cite{zhang2025beyond}, and FiCoCo-V \cite{han2024filter} employ inter-token similarity matrices to identify duplicates. Although versatile, calculating this matrix introduces a time complexity of $O(N^2)$, where $N$ is the number of vision tokens. This computational overhead often offsets the efficiency gains derived from pruning, resulting in suboptimal end-to-end latency.

\begin{figure*}[htbp]
    \centering
    \includegraphics[width=0.8\linewidth]{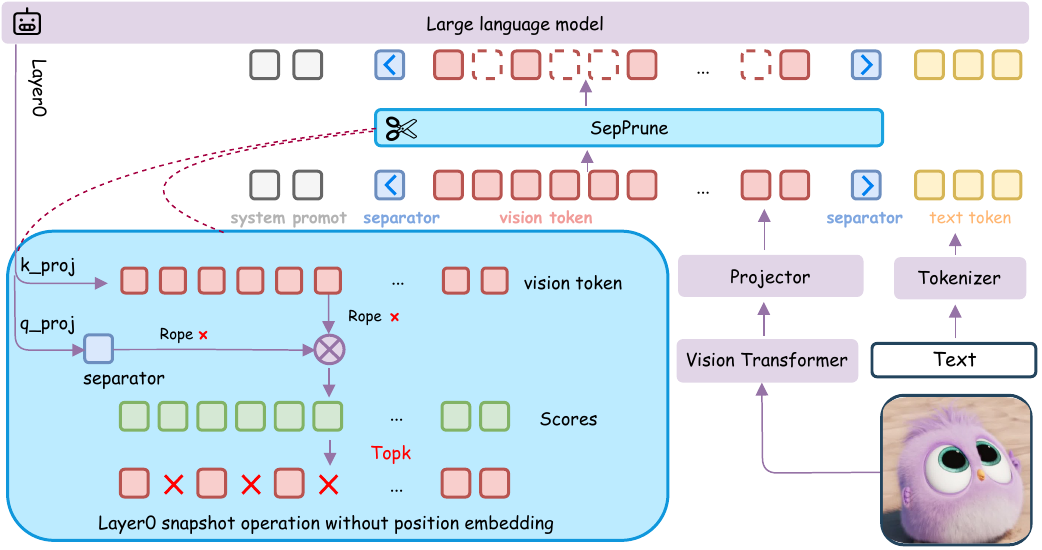}
    \caption{\textbf{Illustration of SepPrune.} We design an efficient vision token selection mechanism for the dynamic vision tokens pruning. This mechanism utilizes  separator token as the Query and the vision tokens as Keys, leveraging the projection parameters from the first layer of the LLM to evaluate the saliency score of each visual token. Notably, no Rotary Position Embeddings (RoPE) operation is introduced into this process. Subsequently, we employ a top-$k$ algorithm to retain only the $k$ most salient vision tokens along with their original positional indices, and this sparsified sequence is then fed into the LLM for further processing. }
    \label{fig:pipeline}
\end{figure*}

\section{Method}
Our analysis shows that the attention scores from both vision and text tokens peak at the modality separator in the shallow layers, while fading in deeper layers. Based on this phenomenon, we hypothesize that the separator acts as a bridge connecting different modalities. Experiments corroborate this hypothesis, as removing the separator significantly impairs the MLLM's performance, thereby demonstrating the necessity of this bridging role.

Inspired by the aforementioned observations, we propose SepPrune. The core methodology of SepPrune involves pruning vision tokens prior to their input into the LLM using the modality separator token. Our method utilizes the modality separator token to perform a unified query and evaluation across all vision tokens. Based on the resulting scores, the vision tokens are ranked, and those with lower scores are subsequently pruned. This approach significantly reduces the computational load on the MLLM during the prefill phase. The architecture of SepPrune is illustrated in Fig. 3.

\subsection{SepPrune: Method Overview} 
SepPrune can be applied immediately after vision tokens are generated. As illustrated in the Fig. 3, the structure of the complete input sequence is as follows: 
\begin{equation}
X_{input} = [T_{sys}, S_{left}, V, S_{right}, T_{usr}],
\end{equation}
where $T_{sys}$ represents the token sequence of the system prompt, $S_{left}$ and $S_{right}$ are the modality separator before and after vision tokens, $V$ is the sequence of vision tokens, and $T_{usr}$ is the sequence of user text tokens. 

To elucidate our method, we first introduce the standard attention mechanism. 
In the standard Transformer attention computation, the Query ($Q$), Key ($K$), and Value matrices are generated from input sequence $X_{input}$ through linear 
projections firstly. In order to perceive the relative position of all tokens, modern 
architectures commonly adopt Rotary Position Embedding (RoPE) \cite{su2023roformerenhancedtransformerrotary}.

\begin{table*}[t]
\centering
\caption{Performance comparison of different pruning methods on Qwen2.5-VL-7B. Score represents the average percentage of performance maintained.}
\label{tab:grouped_final_data}

\footnotesize
\setlength{\tabcolsep}{3pt}

\begin{center}
\begin{tabular}{
  l c c c c c c c c c c c c
}
\toprule
{\textbf{Method}} &{\textbf{ChartQA}} & {\textbf{MME}} & {\textbf{AI2D}} & {\textbf{TextVQA}} & {\textbf{HallBench}} & {\textbf{POPE}} & {\textbf{MMB-EN}} & {\textbf{MMB-CN}} & {\textbf{RealWorldQA}} & {\textbf{SQA}} & {\textbf{Score}} \\
\midrule

\multicolumn{12}{c}{\cellcolor{gray!25}\textbf{Baseline Model}} \\
 Qwen2.5-VL-7B    & 86.1 & 2304 & 80.4 & 84.8 & 46.8 & 86.5 & 82.8 & 83.2 & 68.2 & 83.5 & 100.0\% \\
\midrule

\multicolumn{12}{c}{\cellcolor{gray!25}\textit{ Pruning ratio ($\downarrow$ 60.5\%)}} \\
 FastV (ECCV24)\cite{chen2024image}    & 82.2 & 2317 & 78.8 & 84.1 & 42.4 & 85.2 & 82.0 & 81.8 & 68.1 & 81.5 & 97.7\% \\
 DivPrune (CVPR25)\cite{alvar2025divprune} & 79.6 & 2279 & 78.6 & 81.8 & 43.3 & 84.7 & 81.6 & 82.1 & 67.1 & \textbf{82.4} & 97.0\% \\
 CDPruner (NIPS25)\cite{zhang2025beyond}         & 82.8 & 2327 & 78.9 & 84.2 & 42.5 & 85.4 & 82.2 & \textbf{82.6} & 67.4 & 81.1 & 97.9\% \\
SepPrune           & \textbf{83.7} & \textbf{2335} & \textbf{79.8} & \textbf{84.7} & \textbf{45.0} & \textbf{85.9} & \textbf{82.5} & \textbf{82.6} & \textbf{68.5} & 81.6 & \textbf{99.0\%} \\
\midrule

\multicolumn{12}{c}{\cellcolor{gray!25}\textit{ Pruning ratio ($\downarrow$ 80.2\%)}} \\
 FastV (ECCV24) \cite{chen2024image}    & 70.9 & 2238 & 76.2 & 81.5 & 39.0 & 82.5 & 79.6 & 78.9 & 64.2 & 81.6 & 93.2\% \\
 DivPrune (CVPR25)\cite{alvar2025divprune} & 65.1 & 2184 & 76.5 & 76.0 & 36.4 & 83.7 & 80.0 & 79.6 & 63.9 & 80.7 & 91.2\% \\
 CDPruner (NIPS25) \cite{zhang2025beyond}         & 73.0 & 2245 & 77.5 & 82.4 & 40.1 & \textbf{84.3} & \textbf{80.9} & \textbf{79.9} & 62.0 & 76.9 & 93.6\% \\
SepPrune           & \textbf{75.9} & \textbf{2306} & \textbf{78.2} & \textbf{83.4} & \textbf{43.9} & 83.7 & 80.8 & \textbf{79.9} & \textbf{65.9} & \textbf{81.7} & \textbf{96.3\%} \\
\midrule

\multicolumn{12}{c}{\cellcolor{gray!25}\textit{ Pruning ratio ($\downarrow$ 90.1\%)}} \\
 FastV (ECCV24)\cite{chen2024image}     & 52.2 & 2008 & 71.4 & 73.8 & 33.8 & 74.8 & 72.9 & 72.2 & 56.1 & 79.1 & 83.4\% \\
 DivPrune (CVPR25)\cite{alvar2025divprune} & 50.4 & 2108 & 72.1 & 67.0 & 32.6 & \textbf{81.1} & \textbf{77.8} & \textbf{77.8} & 59.1 & 78.2 & 85.0\% \\
 CDPruner (NIPS25) \cite{zhang2025beyond}         & 59.2 & \textbf{2127} & \textbf{74.0} & 77.8 & \textbf{37.2} & 81.0 & 76.2 & 76.5 & 55.7 & 72.1 & 87.0\% \\
 SepPrune            & \textbf{62.4} & 2045 & 73.5 & \textbf{78.8} & 37.0 & 76.4 & 72.6 & 72.7 & \textbf{60.0} & \textbf{80.1} & \textbf{87.2\%} \\

\bottomrule
\end{tabular}%
\end{center}
\end{table*}

We define this position encoding operation as a function, $\text{RoPE}(\cdot, {pos})$, 
which takes a vector (either a query or a key) and its position, ${pos}$, 
in the sequence as input. This function is applied to each token vector in the 
$Q$ and $K$ matrices, respectively, yielding new position-encoded 
matrices, $Q_{rope}$ and $K_{rope}$. 

Subsequently, the final attention output is computed using these matrices that 
incorporate positional information:

\begin{equation}
\label{eq:rope_qk}
Q_{\text{rope}} = \text{RoPE}(Q, {pos}), \qquad K_{\text{rope}} = \text{RoPE}(K, {pos})
\end{equation}

\begin{equation}
\text{AttnWeights} = \text{softmax}\left(\frac{Q_{\text{rope}} \cdot K_{\text{rope}}^T}{\sqrt{d_k}}\right),
\end{equation}

Here, $d_k$ is the dimension of the query and key vectors. To obtain a metric that effectively evaluates the importance of vision tokens 
at a minimal computational cost, SepPrune performs two key deconstructions 
and simplifications of the standard process described above. 
\subsubsection{Separator as Unified Query}
Based on the previous analysis, we posit that the separator  plays a crucial role in cross-modal interaction. Furthermore, due to the causal attention mask employed during training, only the separator located after the vision tokens $S_{right}$ can access the complete image information. We use this separator as the unified query vector, $Q_{sep}$, which interacts exclusively with the key vectors of all vision tokens, $K_{\text{visual}}$. 
This design stems from a core assumption: as the bridge between the two 
modalities, the separator must have a global perception of 
the entire vision tokens sequence to effectively integrate vision information. Consequently, 
the attention scores between the separator and each vision token can be 
regarded as a global assessment of that token's importance. This approach 
circumvents the substantial overhead of computing a full $N \times N$ attention 
matrix, significantly reducing the computational complexity from $O(N^2)$ to $O(N)$. 
\subsubsection{Removing Positional Encoding for Bias Elimination}
When computing attention scores for pruning purposes, we deliberately omit the application of RoPE function. Although RoPE is crucial for the model's final output, it introduces a positional bias when assessing the intrinsic importance of vision tokens, causing the Query vector to assign higher attention scores to adjacent vision tokens, which leads the pruning scheme to preferentially retain tokens from the bottom of the image while being prone to pruning tokens from the upper regions. Related research  \cite{endo2024featherthrottlerevisitingvisual} also indicates that this local preference phenomenon, induced by position encoding, appears in the shallow layers and diminishes as the model depth increases. To obtain an assessment of importance that is position-agnostic and based purely on content, we remove the position encoding from this calculation.

Integrating these two simplifications, the attention computation formula we use for pruning becomes:
\begin{equation}
Score_{\text{prune}} = \text{softmax}\left(\frac{Q_{sep} K_{\text{visual}}^T}{\sqrt{d_k}}\right),
\end{equation}
where $Score_{\text{prune}}$ is a $1 \times N$ attention score vector, $Q_{sep}$ is the query vector of the separator, and $K_{\text{visual}}$ is the key vector matrix for all vision tokens. 

After obtaining the attention score vector, $Score_{\text{prune}}$, we retain 
the top-$k$ vision tokens with the highest scores. These selected tokens 
constitute the pruned visual sequence, $V'$. Finally, $V'$ is concatenated 
with  separators and text tokens to form the final, shorter input 
sequence that is fed into the LLM.

\section{Experiments}

In this section, we conduct a series of extensive experiments to evaluate the effectiveness of SepPrune . We evaluate SepPrune on Qwen2.5-VL-7B and InternVL3-8B. Furthermore, we have designed detailed ablation studies to provide an in-depth analysis of the contributions of key components in SepPrune and to validate the rationale behind its core mechanism. 

\subsubsection{Evaluation benchmarks}
We conduct experiments on 12 benchmarks covering three key capabilities: (1) \textbf{General Image/Video VQA}: MME, MMBench, ScienceQA, RealWorldQA, VideoMME, and WorldSense; (2) \textbf{Hallucination}: HallBench and POPE; and (3) \textbf{Text-oriented VQA}: TextVQA, OCRBench, ChartQA, and AI2D.

\subsubsection{Comparison methods}
We select two plug-and-play pre-pruning methods, DivPrune and CDPruner, as direct baselines. We also compare against FastV, an internal pruning approach, to validate the efficacy of our non-intrusive method. 

Please refer to the Supplementary Material for detailed experimental settings, as well as the full list of dataset citations and descriptions.

\begin{table*}[t]
\centering
\caption{Performance comparison of different pruning methods on InternVL3-8B. Score represents the average percentage of performance maintained. }
\label{tab:token_pruning_results}

\footnotesize
\setlength{\tabcolsep}{4pt}
\begin{center}
\begin{tabular}{
  l c c c c c c c c c
}
\toprule
\textbf{Method} & \textbf{AI2D} & \textbf{TextVQA} & \textbf{ChartQA} & \textbf{OCRBench} & \textbf{HallBench} & \textbf{MME} & \textbf{MMB-EN} & \textbf{MMB-CN} & \textbf{Score} \\
\midrule
\multicolumn{10}{c}{\cellcolor{gray!25}\textit{Baseline Model}} \\
\ InternVL3-8B & 85.2 & 81.5 & 85.1 & 853 & 50.0 & 2394 & 83.9 & 82.6 & 100.0\% \\
\midrule

\multicolumn{10}{c}{\cellcolor{gray!25}\textit{Retain 256 Tokens ($\downarrow$ 80.0\%)}} \\
\ FastV (ECCV24) \cite{chen2024image}     & 82.2 & 74.4 & 70.7 & 632 & 48.5 & 2348 & \textbf{83.6} & \textbf{82.0} & 92.4\% \\
\ DivPrune (CVPR25)\cite{alvar2025divprune}   & 80.9 & 64.7 & 57.5 & 477 & 38.7 & 2249 & 80.8 & 80.2 & 82.8\% \\
\ CDPruner (NIPS25)\cite{zhang2025beyond}     & \textbf{82.7} & 75.7 & 72.0 & 640 & 48.8 & \textbf{2334} & 83.5 & 81.7 & 92.9\% \\
\ SepPrune            & 82.1 & \textbf{77.1} & \textbf{73.0} & \textbf{646} & \textbf{48.9} & 2332 & 82.5 & 81.4 & \textbf{93.1\%} \\
\midrule

\multicolumn{10}{c}{\cellcolor{gray!25}\textit{Retain 128 Tokens ($\downarrow$ 90.0\%)}} \\
\ FastV (ECCV24)\cite{chen2024image}      & 77.3 & 63.7 & 46.9 & 426 & 42.5 & 2250 & 81.3 & 80.2 & 80.9\% \\
\ DivPrune (CVPR25) \cite{alvar2025divprune}  & 76.4 & 55.6 & 42.7 & 378 & 37.7 & 2166 & 78.4 & 77.6 & 75.7\% \\
\ CDPruner (NIPS25)\cite{zhang2025beyond}     & \textbf{79.9} & 67.5 & 50.8 & \textbf{471} & \textbf{44.6} & \textbf{2282} & \textbf{82.1} & 80.3 & \textbf{83.9\%} \\
\ SepPrune            & 76.4 & \textbf{69.4} & \textbf{52.6} & 466 & 44.3 & 2166 & 81.0 & \textbf{80.3} & 83.0\% \\
\bottomrule
\end{tabular}%
\end{center}
\end{table*}

\subsection{Main results}
 
We evaluate SepPrune on the Qwen2.5-VL-7B model across comprehensive image and video benchmarks. 
\textbf{On image tasks (Table \ref{tab:grouped_final_data})}, SepPrune consistently outperforms state-of-the-art methods (FastV, DivPrune, CDPruner) across all pruning ratios. Notably, at a 60.5\% ratio, it retains 99.0\% of the original performance. This advantage widens at 80.2\% pruning, where SepPrune achieves an overall score of 96.3\%, surpassing CDPruner by nearly 3 percentage points. Even under extreme compression (90.1\%), our method exhibits superior robustness, particularly in information-dense tasks like TextVQA and ChartQA. 

\textbf{On video benchmarks (Table \ref{tab:pruning_performance_comparison})}, SepPrune achieves the best performance on VideoMME and WorldSense at the 0.6 ratio and maintains a leading position on VideoMME (52.0) at the 0.8 ratio. This indicates that SepPrune effectively preserves critical spatiotemporal cues while filtering redundant vision tokens.

To validate \textbf{generalizability}, we applied SepPrune to InternVL3-8B (Table \ref{tab:token_pruning_results}). At 80\% pruning, it significantly outperforms all baselines with a 93.1\% relative score. While CDPruner shows marginal gains in overall score at the extreme 90\% ratio, SepPrune remains dominant in complex reasoning tasks, confirming its adaptability and effectiveness in preserving core semantics across different MLLM architectures.

\begin{table}[h]
\centering
\caption{Performance comparison on VideoMME and WorldSense benchmarks across different pruning ratios.}
\label{tab:pruning_performance_comparison}
\renewcommand{\arraystretch}{1.1} 
\setlength{\tabcolsep}{0pt}       
\footnotesize                     

\begin{tabular*}{\linewidth}{@{\extracolsep{\fill}} l cccccc }
\toprule
& \multicolumn{3}{c}{\textbf{VideoMME}} & \multicolumn{3}{c}{\textbf{WorldSense}} \\
\cmidrule(lr){2-4} \cmidrule(lr){5-7}
\textbf{Method} & \textbf{0.6} & \textbf{0.8} & \textbf{0.9} & \textbf{0.6} & \textbf{0.8} & \textbf{0.9} \\
\midrule

\multicolumn{7}{c}{\cellcolor{gray!15}\textit{No Pruning}} \\
Qwen2.5-VL-7B & \multicolumn{3}{c}{53.2} & \multicolumn{3}{c}{35.1} \\

\midrule

\multicolumn{7}{c}{\cellcolor{gray!25}\textit{Pruning Methods}} \\
FastV \cite{chen2024image}          & 52.4 & 50.9 & 50.0 & 34.4 & 33.3 & 32.1 \\
DivPrune \cite{alvar2025divprune}      & 51.8 & 51.0 & 50.0 & 34.3 & \textbf{33.5} & \textbf{32.8} \\
CDPruner \cite{zhang2025beyond}          & 52.1 & 51.1 & \textbf{50.4} & 32.6 & 33.2 & 32.4 \\
\textbf{SepPrune} & \textbf{52.7} & \textbf{52.0} & 50.1 & \textbf{34.5} & 33.4 & 32.6 \\
\bottomrule
\end{tabular*}
\end{table}

\subsection{Efficiency Analysis}
We evaluate the computational efficiency of SepPrune against DivPrune and CDPruner. Since all three methods prune vision tokens prior to the LLM input---ensuring identical downstream FLOPs reductions---we focus specifically on the algorithmic overhead (detailed complexity analysis provided in the Appendix). Theoretically, SepPrune scales \textbf{linearly} with the number of vision tokens, offering a fundamental advantage over the \textbf{quadratic} complexity of the baselines. This theoretical efficiency is empirically corroborated by MME benchmarks on an NVIDIA H20 GPU (Table \ref{tab:complexity_with_mme_latency}), where SepPrune achieves the lowest end-to-end latency (591.4s) compared to the unpruned baseline (760.4s) and competing methods, validating its superior scalability for high-resolution inputs.

\subsection{Ablation study}
To validate the effectiveness of the core designs in SepPrune, we conducted a series of ablation studies. The experiments were based on the Qwen2.5-VL-7B, with SepPrune achieving a pruning ratio of 80.2\%. 
\subsubsection{Effectiveness of the Separator as a Query}
Due to the constraint of the causal attention mask used during training, only the separator token that appears after the vision token sequence can access the complete image information. Therefore, we use this post-vision separator as the query vector to evaluate and select visual information. 

\begin{table}[t]
\centering
\caption{Effect of pruning 60.5\% vision tokens on model complexity, MME performance, and end-to-end latency for \textbf{Qwen2.5-VL-7B}.}
\label{tab:complexity_with_mme_latency}
\renewcommand{\arraystretch}{1.1} 
\setlength{\tabcolsep}{0pt} 
\footnotesize
\begin{tabular*}{\linewidth}{@{\extracolsep{\fill}} l c c c }
\toprule
\textbf{Method} & \textbf{Time Complexity} & \textbf{MME} & \textbf{Latency} \\
\midrule
\multicolumn{4}{c}{\cellcolor{gray!15}\textit{No Pruning}} \\
Qwen2.5-VL-7B & -- & 2304 & 760.4 \\
\midrule
\multicolumn{4}{c}{\cellcolor{gray!25}\textit{Pruning ratio ($\downarrow$ 60.5)}} \\ 
DivPrune\cite{alvar2025divprune} & $O(N^2 \cdot D)$ & 2279 & 598.7 \\
CDPruner\cite{zhang2025beyond} & $O(N^2 \cdot D + N \cdot M \cdot D)$ & 2327 & 639.9 \\
\textbf{SepPrune} & \textbf{$O(N \cdot D^2)$} & \textbf{2335} & \textbf{591.4} \\
\bottomrule
\end{tabular*}
\end{table}

To verify the superiority of SepPrune, we established two control groups for comparison: (1) using the last token of the LLM's input sequence as the query, and (2) using the separator token that precedes the vision token sequence. The experimental results, as presented in Table \ref{tab:sepprune_ablation_singlecol_no_chartqa}, demonstrate that our chosen query strategy captures the global visual representation more effectively. 
\subsubsection{Effectiveness of the Position-Free Encoding Strategy}
SepPrune removes positional encodings from the Query and Key when computing attention scores. We posit that the positional encodings in the native attention mechanism introduce a positional bias, which can interfere with the model's ability to focus on the content itself. 
To validate this strategy, we compared it against the native Transformer attention mechanism, which includes positional encodings. As shown in Table \ref{tab:sepprune_ablation_singlecol_no_chartqa}, the model's performance improved upon the removal of positional encodings, thereby confirming the rationale of our design.

\begin{table}[htbp]
\centering
\caption{ Ablation studies of SepPrune on Qwen2.5-VL-7B. We analyze the impact of different Query Strategies (using the last token or the starting separator) and the Positional Encoding strategy (with RoPE). }
\label{tab:sepprune_ablation_singlecol_no_chartqa}
\renewcommand{\arraystretch}{1.2}
\setlength{\tabcolsep}{2.5pt} 
\footnotesize 
\begin{tabular}{lccc}
\toprule
\textbf{Method } & \textbf{MME } & \textbf{TextVQA} & \textbf{HallBench } \\
\midrule
\text{SepPrune} & \text{2306} & \text{83.4} & \text{43.9} \\
\midrule
\multicolumn{4}{c}{\cellcolor{gray!25}\textit{Query Strategy}} \\
  Last token    & 2296 ($\downarrow$10) & 82.8 ($\downarrow$0.6) & 42.8 ($\downarrow$1.1) \\
 Starting Separator    & 2292 ($\downarrow$14) & 82.9 ($\downarrow$0.5) & 41.7 ($\downarrow$2.2) \\
\midrule
\multicolumn{4}{c}{\cellcolor{gray!25}\textit{Positional Encoding}} \\
 with RoPE               & 2224 ($\downarrow$82) & 76.2 ($\downarrow$7.2) & 40.0 ($\downarrow$3.9) \\
\bottomrule
\end{tabular}
\end{table}

\section{Conclusion}

In this paper, we propose SepPrune, a plug-and-play method for vision token pruning in MLLMs. Through an in-depth investigation of the modality interaction mechanisms within MLLMs, we reveal the critical role played by the separator token. The attention scores of separator token peak in the shallow layers, bridging the vision tokens and text tokens. Building on this finding, SepPrune innovatively leverages the separator to evaluate and select the most informative vision tokens, all without altering the MLLM architecture. We conducted extensive experiments on Qwen2.5-VL-7B and InternVL3-8B across twelve mainstream multimodal benchmarks. The results demonstrate that SepPrune achieves state-of-the-art  performance.

\section*{Acknowledgment}
This work was in part supported by the National Natural Science Foundation of China under grants 62472399 and Open Fund of APKL of BIIP, IAI, Hefei Comprehensive National Science Center under grants 24YGXT003.

\bibliographystyle{IEEEbib}
\bibliography{main}

\vspace{12pt}

\end{document}